# Automated Information Extraction from Thyroid Operation Narrative: A Comparative Study of GPT-4 and Fine-tuned KoELECTRA

**Dongsuk Jang[†], BS[1, 2], Hyeryun Park[†], MS[1, 2], Jiye Son, MS[1, 2],**
**Hyeonuk Hwang, MD, MS[3, 4], Su-jin Kim, MD, PhD[4], Jinwook Choi, MD, PhD[5]**
[1]Interdisciplinary Program, Bioengineering Major, Seoul National University; [2]Integrated Major in Innovative Medical Science, Seoul National University; [3]Department of Surgery, Ewha Womans University Medical Center; [4]Department of Surgery, Seoul National University Hospital and College of Medicine; [5]Department of Biomedical Engineering, College of Medicine, Seoul National University

**Abstract**

*In the rapidly evolving field of healthcare, the integration of artificial intelligence (AI) has become a pivotal component in the automation of clinical workflows, ushering in a new era of efficiency and accuracy. This study focuses on the transformative capabilities of the fine-tuned KoELECTRA model in comparison to the GPT-4 model, aiming to facilitate automated information extraction from thyroid operation narratives. The current research landscape is dominated by traditional methods heavily reliant on regular expressions, which often face challenges in processing free-style text formats containing critical details of operation records, including frozen biopsy reports. Addressing this, the study leverages advanced natural language processing (NLP) techniques to foster a paradigm shift towards more sophisticated data processing systems. Through this comparative study, we aspire to unveil a more streamlined, precise, and efficient approach to document processing in the healthcare domain, potentially revolutionizing the way medical data is handled and analyzed.*

**Introduction**

In recent years, the healthcare domain has witnessed a profound surge in the exploration and incorporation of AI, including the automation of clinical workflows[1, 2]. Automation of documentation and clinical information extraction is underway, and one of the records that require this process is the operation record. Operation records serve as structured documents that encapsulate intricate details about surgical procedures. Despite their critical role in maintaining transparency and facilitating post-operative care, the process of documenting these records remains laborious and time-consuming. Furthermore, the lapse between the actual surgery and the documentation process can potentially lead to information loss, thereby compromising the quality of the records. This study seeks to address these challenges by leveraging the capabilities of the pre-trained KoELECTRA model[3] to automate the extraction of operational details from free text narratives, particularly focusing on thyroid surgery data. The framework presented in this study marks a pivotal step towards streamlining the documentation process, fostering a seamless transition from a semi-structured format to a more structured and efficient record-keeping system. Through this initiative, we aspire to mitigate the challenges posed by conventional methods, paving the way for a more reliable and efficient documentation process in the healthcare domain.

The necessity for this research stems from the inherent limitations of the existing documentation processes. Traditional methods, reliant on regular expressions, often struggle to accurately process free-style text formats that encompass crucial segments of the operation record such as frozen biopsy reports[4, 5, 6]. By integrating deep learning models into the documentation process, we aim to circumvent these hurdles, facilitating a more nuanced and accurate extraction of information. To the best of our knowledge, this is the inaugural study proposing a method to generate structured operation records utilizing free text, deep learning models, and visualizing surgical outcomes setting a precedent for further advancements in this domain.

In the realm of natural language processing (NLP), there has been a notable paradigm shift marked by the advent of transformer[7] based language models such as BERT[8], and ELECTRA[9]. These innovative models, characterized by their deep learning architectures, have encouraged a wave of research initiatives in NLP, fostering the adaptation of these technologies in a variety of applications. Significantly, they have found extensive utility in named entity recognition (NER). Parallel to these developments, there has been a remarkable surge in the utilization of deep learning models in NER tasks in recent years[10, 11, 12]. These innovative models have shown superior performance in discerning and classifying fragments of named entities in text data, achieving state-of-the-art results[13, 14, 15, 16]. Focusing more narrowly on the Korean language, substantial advancements have been realized with the inception of pre-trained language models tailored for Korean datasets. Initially, the launch of KoBERT[17] signaled a critical leap in this domain, paving the way for further in-depth research into the intricacies of Korean language processing. This momentum has been sustained with the introduction of other pivotal models like KoELECTRA, and more recently, KLUE-RoBERTa[18], fostering a conducive environment for continual research in this specialized segment of language processing.

---

[†]: Equally contributed.

**Methods**

**A. Framework**

We have developed a framework that automates the entire process as shown in Figure 1 of creating structured surgical records from transcript documents concerning thyroid details noted by surgeons. This process is streamlined on a web page where the only requirement is to upload the transcript data which is generated from Naver Clova Note, a Speech-to-Text (STT) platform that records the highest accuracy among several STT platforms[19].

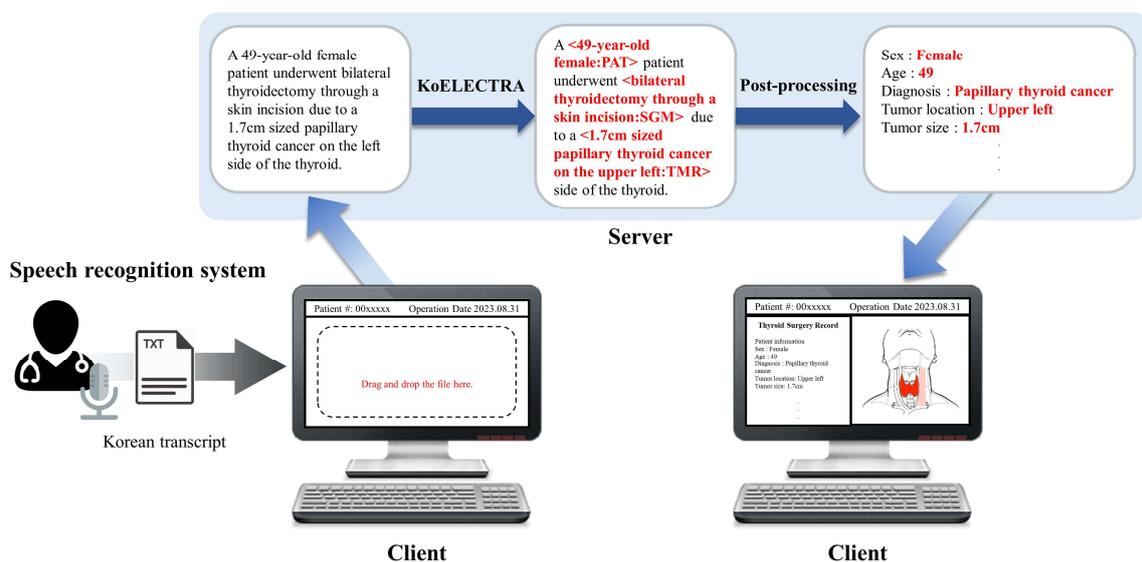

**Figure 1.** Schematic diagram of our framework. As the surgeon records and generates a speech-based transcript, simply upload the transcript from the client's computer to the webpage. Then, the server implementing our fine-tuned KoELECTRA model and post-processing automatically extracts the main information and structures it. Based on the structured result, the client can also generate an image of the main information of the surgery.

The constituent stages of our proposed framework are as follows. The surgeons articulate the specifics of the thyroid surgeries, which are processed through the Naver Clova Note system and transformed into transcripts. Subsequently, the transcript moves into our fine-tuned KoELECTRA model which is intricately trained using a specialized dataset that we exclusively developed. This model plays a pivotal role in identifying and categorizing various pieces of clinical information embedded within the transcript data effectively, marking a significant stride toward achieving structured documentation. As the data progresses through the pipeline, it reaches a post-processing phase where regular expressions are utilized. This stage is instrumental in giving a structured and organized form to the data, streamlining it to be aptly prepared for the creation of the final documentation. It's a pivotal step that ensures the integrity and coherence of the information being documented. Culminating this comprehensive process is the generation of image-based supplementary surgical records. The actual image creation unfolds in a layered approach. Starting with a base layer of blank anatomy as a foundation, we strategically assembled colored fragment layers representing whether the thyroid gland and surrounding regions are resected or preserved, based on the information obtained from the finalized operation records. Stacking these fragment layers onto the base layer allows for a graphical representation of the surgical notes and also offers an intuitive medium to understand and analyze the surgical outcomes in a more engaging and interactive manner.

### B. Dataset

In this study, we have formulated two distinct datasets. One is a dataset built to fine-tune the KoELECTRA model, which has exhibited superior performance in Korean NER, to tailor it to our research needs. The other is a dataset generated through oral recordings by surgeons using existing thyroid surgical records, which is designed to be utilized within the framework we developed.

**B. 1. ThyroNER** We assembled a dataset that contains surgical narratives along with corresponding NER annotations. As depicted in Figure 1, the input data for our framework is a free text consisting of 6-8 sentences detailing a surgical procedure. This data was collected from the surgical records of patients who had undergone thyroid disease surgery at Seoul National University Hospital (SNUH) from January 1, 2020, to June 30, 2021. Initially, thyroid surgeons identified 35 elements essential for thyroid surgical documentation. Subsequently, we outlined 18 broader tags to be extracted during the NER phase. Ultimately, we developed NER-labeled data by the established tagging guidelines. Our dataset encompassed 741 annotated narratives, which were split into 592, 74, and 75 entries for training, validation, and testing datasets respectively.

**B. 2. ThyroTranscript** In an effort to emulate the actual hospital environment and rigorously test our framework, we deliberately incorporated elements akin to real-world situations in the creation of this dataset. During the construction of the audio data, we did not exert stringent control over the recording conditions. For this dataset, surgeons created audio files by orally recording the details of 65 cases of thyroid surgeries conducted at SNUH and also verified the textual records compared to audio files. These narratives simulate various potential complications that might occur during the conversion process, providing a more robust and realistic testing ground for our framework.

### C. Experiments

In this section, we detail the methodology of two distinct experiments in Figure 2, designed to evaluate the effectiveness of our framework. The transcripts used for the experiments were documented in two separate formats.

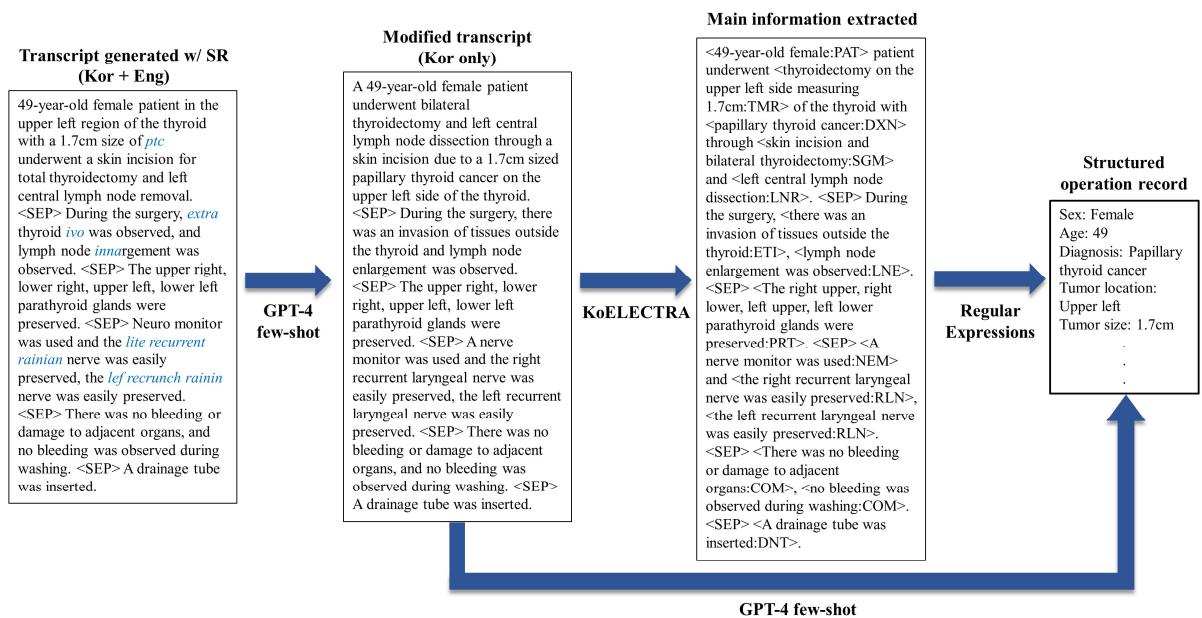

**Figure 2.** Detailed process of experiment with ThyroTranscript mixed language dataset. 'SR' over the table stands for speech recognition, which means converting surgeons' verbal records into transcripts with Naver Clova Note. The process is divided into two parts: One is structuring with KoELECTRA and post-processing and the other is directly structuring with GPT-4 few-shot learning.

The first format contains narratives exclusively in the Korean language, which will be referred to as "Korean only" in the subsequent discussions. The second format includes narratives that are a combination of Korean and English, but have been transliterated into Korean text; this format will be denoted as "Kor + Eng" afterward.

**C. 1. KoELECTRA driven experiment** Transcript serves as the input for our finetuned KoELECTRA model, pre-trained to understand Korean, thereby creating an intermediate text laden with NER annotations. Following this, the intermediate text is subjected to post-processing via regular expressions to tease out major information, thereby morphing it into a structured surgical record format. Based on this final record, an image is generated, encapsulating the essence of the surgery details in a graphical format.

However, when it comes to handling Kor + Eng mixed data in this experiment, the approach is slightly different. Rather than directly feeding the transcript into the KoELECTRA model, it undergoes an initial restructuring. This restructuring is achieved using GPT-4's few-shot learning technique, where it is formatted into a more Korean-centric format before being processed through the KoELECTRA model.

We finetuned KoELECTRA-base-v3 using two GPUs (GeForce RTX 3090). The weight decay was 0, the warm-up proportion was 0.1, the max grad norm was 3, the learning rate was 5e-5, and the training epoch was 40. All results in the tables are obtained using the best epoch weights.

**C. 2. GPT-4 driven experiment** The initial step mirrors the KoELECTRA driven experiment, where audio-recorded files are translated into transcripts through the assistance of the Naver Clova Note. The distinctive feature of this experiment lies in the streamlined approach adopted in the conversion of the transcript to the final structured text. Here, the GPT-4 model, equipped with few-shot learning capabilities, takes the helm, single-handedly transforming the transcript into a structured operation record without the need for an intermediary step. This final text then becomes the basis for generating images that encapsulate the surgical proceedings, maintaining consistency with the final step observed in KoELECTRA driven experiment. This approach represents a more direct pathway to achieve the end goal, demonstrating accuracy in documenting surgical details.

We conducted a few-shot learning with five cases, providing both the transcript and the correct structured record for reference, implementing as many various cases as possible. Moreover, in the prompt instructions, it was specified that if the content concerning any of the 22 classes necessary for the structuring was absent in the transcript, it should be marked as 'not mentioned'. An example of the prompt we used is in Table 1.

**Table 1.** Example of prompt for structuring operation record with GPT-4 experiment and the structured operation record. The original prompt is written in Korean but for readability, we converted it to English in this table.

| Prompt for few-shot learning |
|---|
| Please convert the given document into a structured output format. If the document lacks information for any category, denote that category with 'not mentioned'.<br>Document: """<br>Case 33 A 50-year-old female patient underwent total thyroidectomy and bilateral central lymph node dissection using a skin incision for bilateral thyroid papillary cancer.<br>…<br>A drain was inserted.<br>"""<br>Structured Output: {"Age": 50, "Gender": "Female", "Tumor Location": ["Left", "Right"], "Tumor Size": [1.3, 1.1], … , "Drain Insertion": "Inserted"} |

**Results**

**A. Metric Evaluation of NER**

The results of the KoELECTRA model are presented in Table 2. The model achieves an average macro F1 score of 0.91. For most tags with sufficient data, the F1 scores were above 0.95. However, for some tags with insufficient data such as "RNS", the F1 score was 0.80. The form of content also influences the performance. Whereas tags with a consistent form such as "PAT" and "DNT" were mostly correctly extracted, achieving an F1 score of close to 1, the "SGM" tag showed a relatively low F1 score (0.93) despite the largest number of data. The "ETC" tag with diverse forms and less data was difficult to extract and showed the lowest F1 score (0.50).

**Table 2.** Results of fine-tuned KoELECTRA performances of each tag and macro-average score. "Support" indicates the number of entities per tag in the training set. P, precision; R, recall; F1, f1 score.

| Tag | Details | P | R | F1 | Support |
|---|---|---|---|---|---|
| PAT | Patient demographics (age and gender) | 1.00 | 1.00 | 1.00 | 592 |
| TMR | Tumor location and size before surgery | 0.97 | 1.00 | 0.99 | 567 |
| ATM | Tumor location and size after surgery | 0.98 | 0.98 | 0.98 | 282 |
| DXN | Diagnosis name | 0.98 | 0.98 | 0.98 | 651 |
| LNT | Lymph node transfer, or not | 0.40 | 0.67 | 0.50 | 31 |
| SGM | Surgery method and resection range information | 0.92 | 0.94 | 0.93 | 787 |
| LNR | Lymph node removal, or not | 0.96 | 0.97 | 0.97 | 545 |
| ETI | Invasion information | 0.97 | 0.98 | 0.98 | 513 |
| LNE | Lymph node enlargement, or not | 0.95 | 0.97 | 0.96 | 438 |
| NEM | Using the neural monitor, or not | 1.00 | 1.00 | 1.00 | 537 |
| RLN | Recurrent laryngeal nerve information (preservation, or not and level of difficulty) | 0.99 | 1.00 | 0.99 | 704 |
| SLN | Superior laryngeal nerve information (visual confirmation and preservation, or not) | 1.00 | 0.92 | 0.96 | 117 |
| PRT | Parathyroid information (preservation, or not) | 0.95 | 0.98 | 0.97 | 724 |
| RNS | During resection of lateral cervical lymph node, nerve preservation, or not | 0.67 | 1.00 | 0.80 | 11 |
| COM | Bleeding and damage information | 0.97 | 0.97 | 0.97 | 580 |
| DNT | Drainage tube insertion, or not | 0.97 | 0.99 | 0.98 | 577 |
| FZS | During surgery, frozen section biopsy information | 0.95 | 0.95 | 0.95 | 367 |
| ETC | Other information to record | 0.46 | 0.55 | 0.50 | 72 |
| **Macro** | | **0.89** | **0.94** | **0.91** | **8095** |

**B. Evaluation on structuring operation record**

We evaluated the accuracy of the final structured surgical records generated through the entire framework on 65 cases for each of the 22 classes, calculating the mean accuracy as in Table 3. The assessment was conducted across four distinct experiments characterized by the combinations of language used, either Korean only or Kor + Eng, and the models employed, either the fine-tuned KoELECTRA or the GPT-4 model.

**Table 3.** Accuracy (%) by class with 4 different experiment settings.

| Class | (Korean only) + KoELECTRA | (Kor + Eng) + KoELECTRA | (Korean only) + GPT-4 | (Kor + Eng) + GPT-4 |
|---|---|---|---|---|
| Age | 100 | 100 | 100 | 100 |
| Sex | 100 | 100 | 100 | 100 |
| Tumor location | 100 | 98.39 | 98.46 | 100 |
| Tumor size | 96.92 | 98.39 | 100 | 96.92 |
| Diagnosis name | 98.46 | 95.16 | 100 | 83.08 |
| Surgery method | 98.46 | 66.13 | 100 | 100 |
| Thyroid resection range | 100 | 98.39 | 100 | 98.46 |
| Lymph node removal, or not | 83.08 | 72.58 | 98.46 | 93.85 |
| Capsular invasion, or not | 100 | 100 | 96.92 | 100 |
| Extrathyroidal invasion, or not | 96.92 | 98.39 | 100 | 95.38 |
| Lymph node enlargement | 100 | 62.90 | 100 | 61.54 |

**Table 3. (continued)**

| | | | | |
|---|---|---|---|---|
| Parathyroid preservation status (upper right) | 98.46 | 98.39 | 100 | 100 |
| Parathyroid preservation status (lower right) | 98.46 | 96.77 | 98.46 | 100 |
| Parathyroid preservation status (upper left) | 98.46 | 93.55 | 100 | 98.46 |
| Parathyroid preservation status (lower left) | 98.46 | 93.55 | 100 | 98.46 |
| Use of neural monitor | 98.46 | 98.39 | 100 | 100 |
| Right recurrent laryngeal nerve preservation, or not | 100 | 98.39 | 100 | 100 |
| Left recurrent laryngeal nerve preservation, or not | 95.38 | 98.39 | 100 | 100 |
| Superior laryngeal nerve (right) | 100 | 98.39 | 100 | 100 |
| Superior laryngeal nerve (left) | 100 | 96.77 | 100 | 100 |
| Bleeding when cleaning the surgical site | 96.92 | 98.39 | 100 | 100 |
| Drainage tube insertion, or not | 98.46 | 100 | 100 | 100 |
| **Average** | **98.04** | **93.70** | **99.65** | **96.64** |

In the case of Korean-only data, the fine-tuned KoELECTRA and GPT-4 achieved accuracies of 98.04% and 99.65% respectively. Meanwhile, for the mixed-language dataset, the recorded accuracies were 93.70% for fine-tuned KoELECTRA and 96.64% for GPT-4.

**Discussion**

In this study, we have formulated a pioneering framework that facilitates the seamless transformation of raw verbal recording into structured surgical records through an intricate series of steps integrating Naver Clova Note for STT, KoELECTRA for NER, and regular expression methods for post-processing, thus automating the documentation process in thyroid surgeries. While our results underscore the promising efficiency and accuracy of this model, particularly illustrating KoELECTRA's considerable competence in Korean NER, there are several aspects that warrant further exploration and improvements.

Our study denoted remarkable accuracies of 98.04% and 99.65% with fine-tuned KoELECTRA and GPT-4 respectively when handling Korean-only data. This elucidates that both models exhibit substantial proficiency and that the discrepancy in performance is not significant compared with mixed language data. Interestingly, the accuracies witnessed a decline when processing mixed language data, with fine-tuned KoELECTRA recording 93.70%, and GPT-4 96.64%.

Examining the data in greater depth, we identified specific issues with several classes among 22 total classes, such as 'Diagnosis name', 'Surgery method', 'Lymph node removal, or not', and 'Lymph node enlargement'. In the case of the 'Surgery method', the reduced performance in the mixed language setup with KoELECTRA, manifesting as an accuracy of 66.13%, can be traced back to discrepancies in the terminology used during the experimentation phase. Originally, the ThyroNER dataset utilized for KoELECTRA fine-tuning was composed solely of Korean and designated the 'Surgery method' class as 'skin incision'. However, in the actual experiments where mixed language transcripts were used as inputs, the 'Surgery method' class was referred to as 'open total thyroidectomy' instead of 'skin incision'. This linguistic inconsistency caused the decreased accuracy observed for this class, as it hindered the model's ability to accurately tag the 'Surgery method' class. Moreover, the performance dip in the 'Lymph node removal, or not' class originates from the absence of instances where phrases like 'lymph node dissection was not performed' were not present in the dataset, also creating a learning gap for KoELECTRA model with exception cases. The 'Lymph node enlargement' class also observed lower accuracies in both KoELECTRA and GPT-4 experiments with mixed language data recording 62.90% and 61.54% respectively. This could be traced back to the transcription process, where physicians either grouped descriptors with other surgical information, leading to inaccurate conversions during the STT phase, or instances where the descriptors were lost during STT phase. This indicates potential areas of vulnerability where the STT process may falter in maintaining the accuracy of data transformation. Further, our experiments with Kor + Eng dataset and GPT-4 highlighted unique challenges with the 'Diagnosis name' class, recording 83.08%. The lower accuracy, particularly noticeable when compared to other setups, was attributed to errors occurring during the transformation of initially mixed language transcripts into Korean-only forms via the GPT-4 few-shot. As evidenced in Figure 2, the transformation process inadvertently led to incorrect diagnosis names, showing the areas where the model could potentially stumble.

Our study encounters several limitations that merit attention. Firstly, our framework partially relies on an existing Speech-to-Text (STT) platform during its initial phase. Transitioning to a fully integrated workflow is crucial for more effective and practical use in clinical settings, enhancing overall efficiency and applicability.

Secondly, while KoELECTRA shows notable proficiency in handling Korean language tasks, particularly in Named Entity Recognition (NER), it faces challenges when processing mixed language data. This issue necessitates an additional, complex step where the transcript, initially obtained from STT, must be reconverted into a comprehensive Korean format prior to undergoing KoELECTRA processing. This extra step introduces a degree of complexity, potentially hindering the seamless flow of the data processing.

Furthermore, our research is constrained by the use of a relatively small dataset. This limitation may impede our model's ability to fully represent the vast diversity of real-world scenarios. To mitigate this, we recognize the need for either data augmentation or the acquisition of more extensive data. Such measures are essential for the creation of more diverse datasets, as training with a larger and varied dataset is crucial for improved real-world applicability.

As we forge ahead, our future endeavors are slated to encompass a meticulous evaluation of satisfaction levels among both surgeons and patients within real clinical settings. These evaluations aim to provide invaluable insights into the practical viability and efficiency of our framework, thereby creating a pathway to adjust and adapt the system more robustly, aligning with the dynamic requirements of a clinical environment. We are also optimistic about extending the framework's applicability to include a broader spectrum of departments within hospital setups. This expansion signifies a decisive step towards a comprehensive digital transformation of clinical documentation processes, opening up new avenues for fostering interdisciplinary collaborations and research, thereby elevating the standards of healthcare delivery.

**Available materials**

We have developed and implemented our framework into the 'ThyroDoc' webpage, where users, especially surgeons, can personally upload transcript files and generate structured operation records and image. A demo video illustrating this process can be found at https://github.com/JamesJang26/2024_AMIA. As of now, the webpage server is operational exclusively within the SNUH internal network, hence access from external networks is restricted. The demo video serves as a representation for those unable to access the server directly.

**Conclusion**

This study represents a significant stride towards automating surgical documentation, thereby promising to revolutionize the manner in which surgical records are maintained and utilized. This comprehensive framework not only ensures a seamless transition from raw, unstructured data to a well-organized, structured surgical record, enhancing both efficiency and accuracy in documenting surgical details but also opens avenues in the integration for image generation and analysis. With further research and refinement, this framework has the potential to become an indispensable tool in surgical settings, aiding in the efficient and accurate documentation of surgical procedures. The insights gained from this study will undoubtedly guide future research in this domain, steering it toward the development of more sophisticated and efficient solutions for surgical documentation.

**Ethical considerations**

In compliance with South Korea's 'Personal Information Protection Act', our research necessitated de-identification of personal information. The dataset used in this study lacked any direct identifiers like names, birth dates, or national IDs, and contained only gender, age, and information about thyroid surgery. This falls under quasi-identifiers, which can potentially identify individuals when combined with other information. Our approach aligns with South Korea laws, employing data deletion from numerous classes in the original records, retaining only 22 key classes for use. This ensures adequate anonymization, adhering to the legal requirements.

Regarding the use of OpenAI API for processing surgical record data, we have adhered to OpenAI's data policy. This policy explicitly states that data input into the API for use is not stored, used, or exported by OpenAI. This ensures the privacy and confidentiality of the data utilized in our research are maintained according to OpenAI's guidelines which are possible at https://openai.com/enterprise-privacy.

**Acknowledgments**

This research was conducted in accordance with ethical guidelines and received approval from the Institutional Review Board (IRB) at Seoul National University Hospital. The IRB approval number for this study is IRB No: (2206-174-1336). This research was supported by the Seoul National University Hospital project "Development of a Web-based thyroid surgery record using voice recognition and deep learning (0320222240)", and the Seoul National University College of Medicine initiative "Building voice big data and developing artificial intelligence algorithm for creating voice-based surgical records (800-20200292)", funded by the respective institutions, Republic of Korea.